\pdfoutput=1

\documentclass[11pt]{article}

\usepackage{acl}

\usepackage{times}
\usepackage{latexsym}

\usepackage[T1]{fontenc}

\usepackage[utf8]{inputenc}

\usepackage{times}
\usepackage{latexsym}
\usepackage{graphicx}
\usepackage{arydshln}

\usepackage{multirow}
\usepackage{multicol}
\usepackage{algorithm}
\usepackage[noend]{algpseudocode}
\usepackage{microtype}
\usepackage{amsmath,amssymb,bm}
\usepackage{tabularx}
\usepackage{booktabs,arydshln}
\usepackage{tabu}

\title{\textsc{Subs}: Subtree Substitution for Compositional Semantic Parsing}

\author{Jingfeng Yang$^{\dagger\star}$ \quad Le Zhang$^{\ddagger\star}$  \quad Diyi Yang$^{\dagger}$\\
  $^{\dagger}$ Georgia Institute of Technology\\
  $^{\ddagger}$ Fudan University\\
  {\tt jingfengyangpku@gmail.com \quad zhangle18@fudan.edu.cn}\\
  {\tt dyang888@gatech.edu } \\
  }

\begin{document}
\maketitle
\begin{abstract}
Although sequence-to-sequence models often achieve good performance in semantic parsing for i.i.d. data, their performance is still inferior in compositional generalization. Several data augmentation methods have been proposed to alleviate this problem. However, prior work only leveraged  superficial grammar or rules for data augmentation, which resulted in limited improvement. We propose to use subtree substitution for compositional data augmentation, where we consider subtrees with similar semantic functions as exchangeable. Our experiments showed that such augmented data led to significantly better performance on \textsc{Scan} and \textsc{GeoQuery}, and reached new SOTA on compositional split of \textsc{GeoQuery}. We have publicly released our
code at \url{https://github.com/GT-SALT/SUBS} .
\end{abstract}

\section{Introduction}

\let\thefootnote\relax
\footnotetext{$\star$Equal Contribution. Jingfeng Yang proposed subtree substitution data augmentation for compositional semantic parsing, implemented augmentation and LSTM/BART parsers, and ran \textsc{Scfg}/\textsc{Geca} baselines. Le Zhang induced span trees and ran span-based semantic parsing baseline.}
\let\thefootnote\svthefootnote

Semantic parsing transforms natural language utterances to formal language. Because meaning representations or programs are essentially compositional, semantic parsing is an ideal testbed for compositional generalization. Although neural seq2seq models could achieve state-of-the-art performance in semantic parsing for i.i.d. data, they failed at compositional generalization due to lack of reasoning ability. That is, they do not generalize well to formal language structures that were not seen at training time. For example, a model that observes at training time the questions ``\textit{What is the population of the largest state?}'' and ``\textit{What is the largest city in USA?}'' may fail to generalize to questions such as ``\textit{What is the population of the largest city in USA?}''. This leads to large performance drops on data splits designed to measure compositional generalization (compositional splits), in contrast to the generalization abilities of humans. 

To improve compositional generalization in semantic parsing (compositional semantic parsing), prior work focused on incorporating inductive biases directly to models or data augmentation. From the model perspective, some work used neural-symbolic models \cite{chen2020compositional}, generated intermediate discrete structures \cite{herzig2020span, zheng2020compositional}, or conducted meta-learning \cite{lake2019compositional}. From the data perspective, \citet{jia2016data} proposed to recombine data with simple synchronous context-free grammar (\textsc{Scfg}), despite not for compositional generalization.  \citet{andreas2019good} used some simple rules for data augmentation, where tokens with the same context were considered as exchangeable. 
Such techniques are still limited 
since they only leveraged superficial grammars or rules, and failed when there are linguistically rich phrases or clauses. 
\begin{figure*}[t]
\begin{center}
\includegraphics[width=0.8\linewidth]{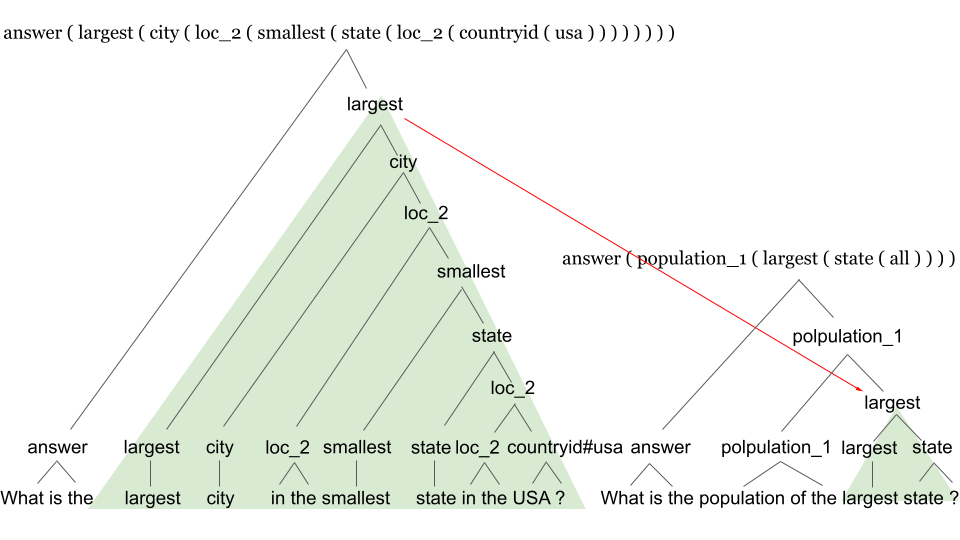}
\end{center}
\caption{\label{fig:annotation}Subtree substitution results in an augmented example. Natural Language: \textit{What is the population of the largest city in the smallest state in the USA ?} Formal Language: \texttt{answer ( population\_1 ( largest ( city ( loc\_2 ( smallest ( state ( loc\_2 ( countryid ( usa ) ) ) ) ) ) ) ) )}.
}\vspace{-0.12in}
\end{figure*}

To fill this gap, we propose to augment the training data of semantic parsing with diverse compositional examples 
based on induced or annotated (semantic and syntactic) trees.
 Specifically, we propose to exchange subtrees where roots have similar meaning functions. Since we consider all hierarchies in all trees, deep structures and complex phrases or clauses are considered for data augmentation, which is key for compositional generalization. For instance, in Figure \ref{fig:annotation}, if we exchange subtrees with ``\textit{largest}'' as meaning function of its root, composition of ``\textit{population of the}'' and ``\textit{largest city in the smallest state in the USA}'' results in a new augmented structure ``\textit{population of the largest city in the smallest state in the USA}''. Although certain substructure substitution methods were explored in other NLP tasks \cite{shi2021substructure}, subtree substitution with fine-grained meaning functions has been under-explored. 
Our experiments showed that such augmented data led to significantly better performance on \textsc{Scan} \cite{lake2018generalization} and \textsc{GeoQuery}, and reached new SOTA on compositional split of \textsc{GeoQuery}.

\section{Methods}

\paragraph{Span trees}
Suppose training set is $\{(x^i,z^i)\}_{i=1}^N$, where  $x_i$ is a natural language utterance and $z_i$ is the corresponding program.
An utterance $x$ can be mapped to a span tree $T$, such that program($T$)= $z$, where the deterministic function program($\cdot$) maps span trees to programs \cite{herzig2020span}.

As shown in Figure \ref{fig:annotation}, a span tree $T$ is a tree where each node covers a span ($i$, $j$) with tokens $x_{i:j}$ = ($x_i, x_{i+1}, \cdots , x_j$). A span subtree can be viewed as a mapping from every span ($i$, $j$) to a single category $c \in C$, where $C$ is a set of domain-specific categories representing domain constants, which include entities (e.g. \textsl{countryid\#usa} in Figure \ref{fig:annotation}) and predicates (e.g. \textsl{loc\_2} in Figure \ref{fig:annotation}). The final program can be computed from the span tree deterministically by the function program(·). Concretely, program($T$) iterates over the nodes in $T$ bottom-up, and generates a program $z_{i:j}$ for each node covering the span ($i$, $j$). For a terminal node, $z_{i:j} = c$. For an internal node, $z_{i:j}$ is determined by composing the programs of its children, $z_{i:s}$ and $z_{s:j}$ where $s$ is the split point. As in Combinatory Categorical Grammar, composition is simply function application, where a domain-specific type system is used to determine which child is the function and which is the argument.
Span trees can be induced by a hard-EM algorithm or semi-automatically annotated. We refer the reader to \citet{herzig2020span} to see how to obtain span-trees.

\subsection{Subtree Substitution (\textsc{Subs})}

As shown in Figure \ref{fig:annotation}, we consider span subtrees with similar semantic functions as exchangeable. Formally, func(·) maps a subprogram to a semantic category, and subtrees with the same semantic categories have similar semantic functions. For two data points ($x^1, z^1$) and ($x^2, z^2$), if func($z^1_{i_1:j_1}$) = func($z^2_{i_2:j_2}$), we obtain a new augmented  ($x', z'$):
\[x' = x^1_{:i_1} + x^2_{i_2:j_2} + x^1_{j_1:},  z'=z^1\backslash z^1_{i_1:j_1} / z^2_{i_2:j_2}\]
Definition of func(·) may vary in different dataset. One straightforward way is to extract the outside predicate in $z_{i:j}$ as its semantic category, which is used on \textsc{GeoQuery}, such as  func(\texttt{largest ( state ( all ) ) )}) = \texttt{largest}. 





\subsection{Semantic Parsing}

After getting augmented data by subtree substitution, we then combine augmented data and the original training data to train a seq2seq semantic parser, where we choose LSTM models with attention \cite{luong2015effective} and copying mechanism \cite{gu2016incorporating}, or pretrained $\text{BART}_{\text{Large}}$ \cite{lewis-etal-2020-bart} as the seq2seq model architecture.

\begin{table}
\centering

\setlength\tabcolsep{4pt}
\begin{tabular}{l|c|l}\toprule
                            & \textsc{Right} & \textsc{AroundRight}  \\\midrule
LSTM                  & 0.00           & 1.00  (2800 updates)       \\
LSTM  + $\textsc{Subs}$                                         & 1.00           & 1.00  (800 updates)  \\\bottomrule
\end{tabular}
\caption{\label{tab:widgets1} Accuracy of diagnostic experiments on \textsc{Scan}.}\vspace{-0.1in}
\end{table}


\begin{table}
\centering

\setlength\tabcolsep{4pt}
\begin{tabular}{l|c|c}\toprule
                            & Question & Query  \\\midrule
\citet{herzig2020span} & 0.78 & 0.75\\\midrule

LSTM                  & 0.75           & 0.58         \\
+ \textsc{Scfg}  (Jia et al., 2016)                                  & 0.80           & 0.68         \\
+ \textsc{Geca}   (Andreas, 2019)                                          & 0.77           & 0.60         \\
+ $\textsc{Subs}$ (ours, induced tree)        & 0.79         & \textbf{0.72}  \\
+ $\textsc{Subs}$ (ours, gold tree)        & \textbf{0.81}         & \textbf{0.79}       \\\midrule
BART                                             & 0.91         & 0.85         \\
+ $\textsc{Subs}$ (ours, induced tree)  & 0.91       & 0.85 \\
+ $\textsc{Subs}$ (ours, gold tree)  & \textbf{0.93}       & \textbf{0.88} \\\bottomrule
\end{tabular}
\caption{\label{tab:widgets13} Exact-match accuracy on i.i.d. (Question) and compositional (Query) splits of \textsc{GeoQuery} dataset. }\vspace{-0.1in}
\end{table}


\section{Experiments and Results}
\textbf{Dataset}
We first use \textsc{Scan} \cite{lake2018generalization} as a diagnostic dataset to test the performance of subtree substitution in compositional semantic parsing. \textsc{Scan} is a synthetic dataset, which consists of simple English commands paired with sequences of discrete actions. We use the program version of \citet{herzig2020span}. For instance, ``\textit{run right after jump}'' corresponds to the program ``\texttt{i\_after ( i\_run ( i\_right ) , i\_jump )}''. Also, semi-automatically annotated span trees from \citet{herzig2020span} are used for subtree substitution. To test compositional semantic parsing, we use the \textit{Primitive right} (\textsc{Right}) and \textit{Primitive around right} (\textsc{AroundRight}) compositional splits from \citet{loula2018rearranging}, where templates of the form \textit{Primitive right} and \textit{Primitive around right} (respectively) appear only in the test set. In these templates, \textit{Primitive} stands for \textit{jump}, \textit{walk}, \textit{run}, or \textit{look}. For simplicity, func(·) is defined only on \texttt{i\_right} and \texttt{i\_left}, where func(\texttt{i\_right}) = func(\texttt{i\_left}) = \texttt{direction}. That is, all ``\texttt{i\_right}'' and ``\texttt{i\_left}'' appear as leaf nodes in span trees and they are exchangeable. 

We use \textsc{GeoQuery} dataset to test the performance of subtree substitution in both i.i.d. and compositional generalization for semantic parsing. \textsc{GeoQuery} contains 880 questions about US
geography \cite{zelle1996learning}. Following \citet{herzig2020span}, we use the variable-free FunQL formalism from \citet{kate2005learning}. The i.i.d. split (Question), which is randomly sampled from the whole dataset, contains 513/57/256 instances for train/dev/test set. The compositional split (Query) contains 519/54/253 examples for train/dev/test set, where templates created by anonymizing entities are used to split the dataset, to make sure that all examples sharing a template are assigned to the same set \cite{finegan2018improving}. As for span trees, we use semi-automatically annotated span trees (gold tree) released by \citet{herzig2020span}. Alternatively, we use the span trees induced by \citet{herzig2020span}'s span-based semantic parsing, without any human labour.

\begin{table*}
\centering
\small
\setlength\tabcolsep{4pt}
\begin{tabular}{c|c|c|c|c|c|c}\toprule
 & training instances & augmented instances    & avg att l        & max att l          & avg prg l     & max prg l          \\\midrule
\textsc{Geca}                & 519 & 804          & 8.85                   & 18             & 15.96                             & 29                               \\
$\textsc{Subs}$  & 519 & 29039       & 10.43                    & 26             & 19.33                           & 43                         \\\midrule
& avg seg l & max seg l & avg att seg l & max att seg l & avg prg seg l & max prg seg l \\\midrule
\textsc{Geca}                 & 1.93         & 4                                 & -                  & -                  & -                   & -                   \\
$\textsc{Subs}$ & 5.99         & 25                                & 3.98               & 13                 & 8.01                & 25       \\\bottomrule          
\end{tabular}
\caption{Complexity of augmented examples on the Query split of \textsc{GeoQuery} dataset, which is measured by maximal (max) and average (avg) lengths (l) of exchanged segments (seg) and resulted utterances(att)/programs(prg). }\label{tab:widgets2}\vspace{-0.1in}
\end{table*}

\begin{table}
\centering

\setlength\tabcolsep{4pt}
\begin{tabular}{l|c|c|c|c}\toprule
                            & 50 & 100 & 200 & 519 \\\midrule
BART                  & 0.64         & 0.72 & 0.79 & 0.85     \\
BART  + $\textsc{Subs}$          & \textbf{0.67}        & \textbf{0.79} & \textbf{0.85} & \textbf{0.88}   \\\bottomrule
\end{tabular}
\caption{\label{tab:widgets5} \small{Effect of numbers of training examples on compositional split of \textsc{GeoQuery}.}}\vspace{-0.1in}
\end{table}

\subsection{Diagnostic Results}

Results of diagnostic experiments on \textsc{Scan} dataset are shown in Table \ref{tab:widgets1}, where we use LSTM parser without data augmentation as the baseline. 
We can see that on the \textsc{Right} split, LSTM seq2seq semantic parser could only achieve zero exact-match accuracy without any data augmentation techniques, which means that the model's compositional generalizibility on the \textsc{Right} split is very poor. After adding our augmented data with subtree substitution, we achieve an exact-match accuracy of 100\%. Actually, we got 6660 augmented examples besides the original 12180 training examples. Among all augmented examples, 3351 examples are in the test set, which means 74.87\% of 4476 test examples are recovered by subtree substitution. On the \textsc{AroundRight} split, using LSTM seq2seq semantic parser could already achieve 100\% exact-match accuracy, which means that the model learned from \textit{Primitive right} and \textit{Primitive opposite right} generalize to \textit{Primitive around right} well in our program format ``\texttt{i\_primitive ( i\_around ( i\_right ) )}''. After adding our augmented examples, the parser converged to 100\% exact-match accuracy faster, where our method requires around 800 updates to converge while baseline model requires 2800 updates with the same batch size 64. 

\subsection{Main Results}
Table \ref{tab:widgets13} shows the results of experiments on \textsc{GeoQuery} dataset, where we examined both seq2seq LSTM and $\text{BART}_{\text{Large}}$ parsers. LSTM and BART parsers without any data augmentation are simplest baselines. We also compare with other two data augmentation methods as additional baselines, recombining data with simple \textsc{Scfg} \cite{jia2016data}  or using simple rules for Good Enough Data Augmentation (\textsc{Geca})  \cite{andreas2019good}, which were proven useful for compositional semantic parsing. 
We can see that on the Question split, adding augmented data from (gold) subtree substitution leads to improvements for both LSTM and BART seq2seq models, suggesting that subtree substitution as data augmentation helps i.i.d generalization for semantic parsing. 
On  the Query split, (gold) subtree substitution achieves more substantial improvements over seq2seq baseline models (absolute 21\% and 3\% improvements of the exact-match accuracy for LSTM and BART respectively), achieving state-of-the-art results. Moreover, our methods are also better than the two data augmentation baselines. Therefore, subtree substitution is a simple yet effective compositional data augmentation method for compositional semantic parsing. 
With (induced) subtree substitution, \textsc{Subs} still achieves improvements for LSTM models.

\noindent
\textbf{Analysis of Augmented Data}
We further examine why subtree substitution could achieve much better performance by analyzing its augmented data. As shown in Table \ref{tab:widgets2}, \textsc{Geca} only identifies and exchanges very simple structures, where the average and maximal length of exchanged segments are 1.93 and 4.  A closer look at these augmented data shows that nearly all of these segments are simple entities (e.g. \textsc{State}: ``Illinois'', ``Arizona'' etc.) or other Nouns (e.g. ``area'', ``population'' etc.). In contrast, subtree substitution can identify and exchange much more complex structures, where the average and maximal length of exchanged segments are 5.99 and 25. For example, \textit{largest city in the smallest state in the USA} and \textit{largest state} are identified as exchangeable. As a result, subtree substitution could produce more complex utterance and program pairs, where the average and maximal length of these resulted utterances are  10.43 and 26, compared with the average (8.53) and maximal (18) length of utterances returned by \textsc{\textsc{Geca}}.
Moreover, subtree substitution could generate much more augmented instances, because it  can identify more complex structures besides those simple ones identified by \textsc{Geca}. Compared with \textsc{Scfg}, \textsc{Subs} could also identify complex structures automatically with subtrees, while \textsc{Scfg} only handle simple phrases defined by rules.

\noindent
\textbf{Effect of Training Data Size}
Table \ref{tab:widgets5} shows that with more training examples, models' performances improve. In all settings, using (gold) subtree substitution boosts the performance of BART. When there are 100 and 200 training examples, the improvement is more significant, demonstrating the effectiveness of \textsc{Subs} in the few-shot setting.

\section{Related Work}\vspace{-0.06in}
Several data augmentation methods have been introduced for (compositional) semantic parsing. \citet{jia2016data} recombined data by \textsc{Scfg}, and \citet{andreas2019good} used some simple rules to exchange tokens with the same context. However, they leveraged only superficial grammars or rules, which has limited capacity to identify complex structures. \citet{akyurek2020learning} learned to recombine and resample data with a prototype-based generative model, instead of using rules. Certain substructure substitution methods have been explored for data augmentation in other NLP tasks \cite{shi2021substructure}. Dependency tree cropping and rotation within sentence was used in low-resource language POS tagging \cite{csahin2019data} and dependency parsing \cite{vania2019systematic}. Dependency tree swapping was explored in low-resource language dependency parsing \cite{dehouck2020data}, and Universal Dependency features was used for zero-shot cross-lingual semantic parsing \cite{yang2021frustratingly}. However, subtree substitution with fine-grained meaning functions has not been examined. Some rule-based data augmentation methods were also explored in table semantic parsing \cite{eisenschlos2020understanding, yang2022tableformer}. To the best of our knowledge, we are the first to explore tree 
manipulation for semantic parsing.  

\section{Conclusion}\vspace{-0.06in}
This work proposed to use subtree substitution to compositionally augment the data of semantic parsing to help the compositional generalization. Our method achieved significant improvements over seq2seq models, other data augmentation methods and span-based semantic parsing. 

\section*{Acknowledgements} The authors would like to thank reviewers for their helpful insights and feedback. This work is funded in part by a grant from Amazon and Salesforce.

\bibliography{anthology,custom}
\bibliographystyle{acl_natbib}

\appendix

\section{Training Details}

We adapted OpenNMT \cite{klein2017opennmt} for LSTM model with attention and copying mechanism, while used fairseq \cite{ott2019fairseq} to implement BART model.

We manually tune the hyper-parameters. For LSTM models, we use one-layer bidirectional LSTM in the encoder side and one-layer unidirectional LSTM in the decoder side. We use dropout with 0.5 as dropout rate and Adam optimizer with a learning rate of 0.001. We use MLP attention and reuse attention scores as copying scores. On \textsc{GeoQuery}, the batch size is set to 1 sentence without augmented data and set to 64 sentences with augmented data. On \textsc{Scan}, all batch sizes are 64 sentences. For BART models, we use BART large models.  We use Adam as optimizer with a learning rate 1e-5. We use dropout and attention dropout with 0.1 as dropout rate. Also, we use label smoothing with a rate 0.1. Batch sizes are 1024 tokens. Besides, we employ a weight-decay rate 0.01.
All the parameters are manually tuned based on the dev performance.

We train all models on NVIDIA A100 SXM4 40 GB GPU. We set the max training epoch to be 100 and select the best performed epoch according to dev performance. Training process on each clause or whole sequence could be finished within 3 hours.

For baselines with other data augmentation methods, we reran \textsc{Geca} and \textsc{Scfg} on this FunQL formalism of \textsc{GeoQuery} and these splits with annotated span trees. That's why our results are a little different from the reported results in the original paper. We got similar results with their source code and our code on our data, in order to make sure that there is no problem with our results and code. 

We got the same denotation accuracy as reported by \citet{herzig2020span}, but we reported  exact-match accuracy on Table \ref{tab:widgets13} for fair comparison.

\end{document}